\documentclass[10pt,twocolumn,letterpaper]{article}

\usepackage[pagenumbers]{cvpr} %

\usepackage{graphicx}
\usepackage{amsmath}
\usepackage{amsfonts}
\usepackage{bm}
\usepackage{amssymb}
\usepackage{booktabs}
\usepackage{xcolor}
\usepackage{float}

\usepackage[pagebackref,breaklinks,colorlinks]{hyperref}
\usepackage{algorithmic}
\usepackage{algorithm}

\usepackage[capitalize]{cleveref}
\crefname{section}{Sec.}{Secs.}
\Crefname{section}{Section}{Sections}
\Crefname{table}{Table}{Tables}
\crefname{table}{Tab.}{Tabs.}

\def\cvprPaperID{XXXX} %
\def\confName{XXXX}
\def\confYear{2024}

\newcommand{\calb}[1]{\boldsymbol{\mathcal{#1}}}
\newcommand{\bs}[1]{{\boldsymbol{#1}}}

\usepackage{microtype}  %
\usepackage{balance}  %
\usepackage{overpic}
\usepackage{enumitem} %
\usepackage{overpic} %
\usepackage{color}

\definecolor{turquoise}{cmyk}{0.65,0,0.1,0.3}
\definecolor{purple}{rgb}{0.65,0,0.65}
\definecolor{dark_green}{rgb}{0, 0.5, 0}
\definecolor{orange}{rgb}{0.8, 0.6, 0.2}
\definecolor{red}{rgb}{0.8, 0.2, 0.2}
\definecolor{darkred}{rgb}{0.6, 0.1, 0.05}
\definecolor{blueish}{rgb}{0.0, 0.3, .6}
\definecolor{light_gray}{rgb}{0.7, 0.7, .7}
\definecolor{pink}{rgb}{1, 0, 1}
\definecolor{greyblue}{rgb}{0.25, 0.25, 1}

\usepackage{blindtext}

\renewcommand{\paragraph}[1]{\vspace{1em}\noindent\textbf{#1}.}

\begin{document}
\title{Generative Rendering: Controllable 4D-Guided Video Generation with 2D Diffusion Models}

\author{
Shengqu Cai$^{1,2,\triangle}$
\quad Duygu Ceylan$^{2*}$
\quad Matheus Gadelha$^{2*}$
\quad Chun-Hao Paul Huang$^{2}$\\
Tuanfeng Yang Wang$^{2}$
\quad Gordon Wetzstein$^{1}$
\\
$^{1}$Stanford University\quad
$^{2}$Adobe Research
}

\twocolumn[{%
  \renewcommand\twocolumn[1][]{#1}%
\maketitle
\begin{center}
  \newcommand{\teaserwidth}{1.0\textwidth}
  \centerline{
    \includegraphics[width=\teaserwidth]{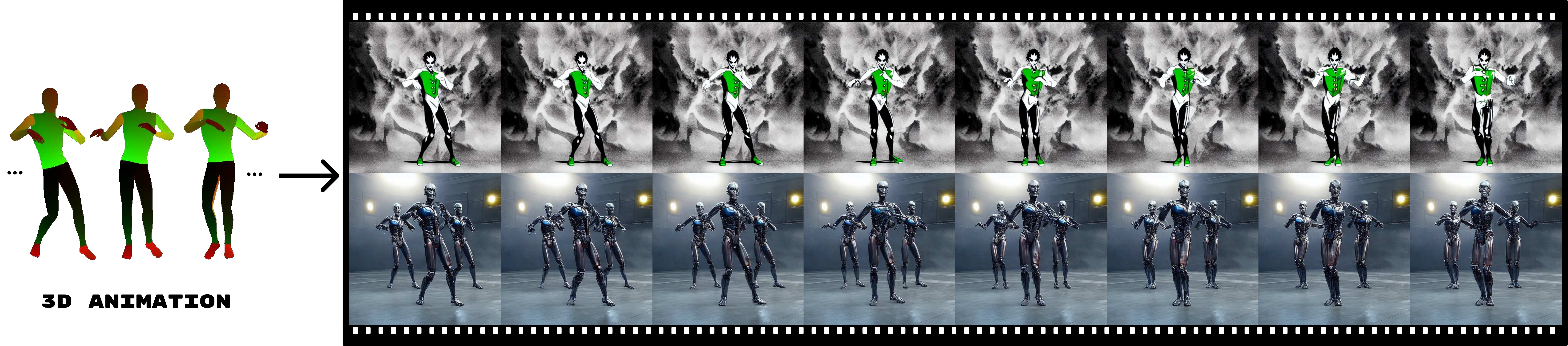}
  }
  \captionsetup{type=figure}
  \captionof{figure}{
    \textbf{Generative Rendering} is a novel diffusion-based approach to render 3D animated but untextured scenes (left) directly to a stylized animation (right), with styles specified by text prompts. Our approach offers new levels of user control to image generation models.
  }
  \vspace{-0.025in}
  \label{fig:teaser}
 \end{center}%
}]

\begin{abstract}
Traditional 3D content creation tools empower users to bring their imagination to life by giving them direct control over a scene's geometry, appearance, motion, and camera path. Creating computer-generated videos, however, is a tedious manual process, which can be automated by emerging text-to-video diffusion models. Despite great promise, video diffusion models are difficult to control, hindering a user to apply their own creativity rather than amplifying it. To address this challenge, we present a novel approach that combines the controllability of dynamic 3D meshes with the expressivity and editability of emerging diffusion models. For this purpose, our approach takes an animated, low-fidelity rendered mesh as input and injects the ground truth correspondence information obtained from the dynamic mesh into various stages of a pre-trained text-to-image generation model to output high-quality and temporally consistent frames. We demonstrate our approach on various examples where motion can be obtained by animating rigged assets or changing the camera path.
Project page: \href{https://primecai.github.io/generative_rendering}{primecai.github.io{\slash}generative\_rendering}.
\end{abstract}
\noindent\let\thefootnote\relax\footnote{$\triangle$: Part of this work was done during an internship at Adobe Research.}
\noindent\let\thefootnote\relax\footnote{* : Equal contribution.}

\section{Introduction}
\label{sec:intro}

Artists, designers, architects, and other creators rely on traditional 3D content creation tools to render computer-generated videos. Unfortunately, existing 3D workflows are laborious, time consuming, and require expertise. Emerging generative artificial intelligence tools, such as text-to-image (T2I) and text-to-video (T2V) models, solve these issues by automating many of the manual steps of traditional workflows. Video generation, however, is difficult to control in that it is not easily possible to specify scene layout and motion in a temporally consistent manner.

Recent approaches have attempted to control diffusion models. For example, ControlNet~\cite{zhang2023controlnet} uses a pre-trained T2I diffusion model and finetunes an adapter network that is conditioned on depth, pose, or edge images to control the layout. This strategy is successful for generating individual frames, but results in flicker for video generation. 
Other approaches aim at learning the complex types of motions encountered in natural videos directly~\cite{ho2022imagenvideo,singer2022makeavideo,li2023videogen,zhou2022magicvideo,wang2023modelscope,he2022latent,yu2023video,blattmann2023alignyourlatents, ge2023pyoco, guo2023animatediff}. While successful in generating smooth motions, these approaches are not easily controllable. Finally, video-to-video diffusion models ~\cite{ceylan2023pix2video, geyer2023tokenflow, khachatryan2023text2videozero, wu2022tuneavideo} enable video editing and stylization, but they require a high-fidelity video as input, which is not always available.

In this paper, we aim to combine the power of 3D workflows with T2I models for generating 4D-guided stylized animations. Specifically, we leverage 3D tools to rapidly prototype proxy geometry and motion of a scene (e.g., camera paths, physically based simulation, or character animation) while utilizing the T2I generation model as a \emph{renderer} to output the final stylized animations (see Fig.~\ref{fig:teaser}).
At the core of our method is the ability to leverage the ground truth 4D spatio-temporal correspondences that can be obtained from an animated 3D scene to guide the image generation. First, we leverage the correspondence information to effectively perform noise initialization which is critical for temporal consistency. Many recent video-to-video stylization methods resort to inverting the source video to obtain consistent noise initialization~\cite{ceylan2023pix2video,geyer2023tokenflow}. However, this is not possible in our setup since we do not have an initial source video and inversion does not work well on untextured renderings of the 3D scene. Instead, we propose to utilize the canonical representation (i.e., UV space) of the 3D scene to initialize random noise which is then projected to each frame of the animation. Next, we enrich the self-attention layers of the image diffusion model which are known to contain spatial appearance information vital for obtaining consistent renderings~\cite{ceylan2023pix2video,wu2022tuneavideo}. Specifically, we propagate both the input and output of self-attention layers across the frames via known correspondences to enforce more consistent results. Finally, similar to previous work~\cite{zhang2023controlnet}, we use the depth cues rendered from the 3D scene in conjunction with control models to provide structure guidance. In summary, our framework treats the pre-trained T2I diffusion model as a multi-frame renderer that is able to capture the correspondences while maintaining high-fidelity and consistent generations.

We evaluate our method on a variety of scenes %
and demonstrate its advantages with comparisons to state-of-the-art methods. In summary, our contributions include:
\begin{itemize}
[leftmargin=*]
\setlength\itemsep{-.1em}
    \item We present a novel framework for 4D-guided animation synthesis utilizing the pre-trained T2I generation models as a multi-frame renderer. %
    \item We enhance the self-attention layers of the image generation model by performing correspondence-aware blending of both input and output features to enforce consistent appearance synthesis.
    \item We introduce a UV-space noise initialization mechanism. Combined with the correspondence-aware attention mechanism, this enables better consistency across frame generations.
\end{itemize}
\begin{figure*}
\begin{center}
\centering
\includegraphics[width=0.99\linewidth]{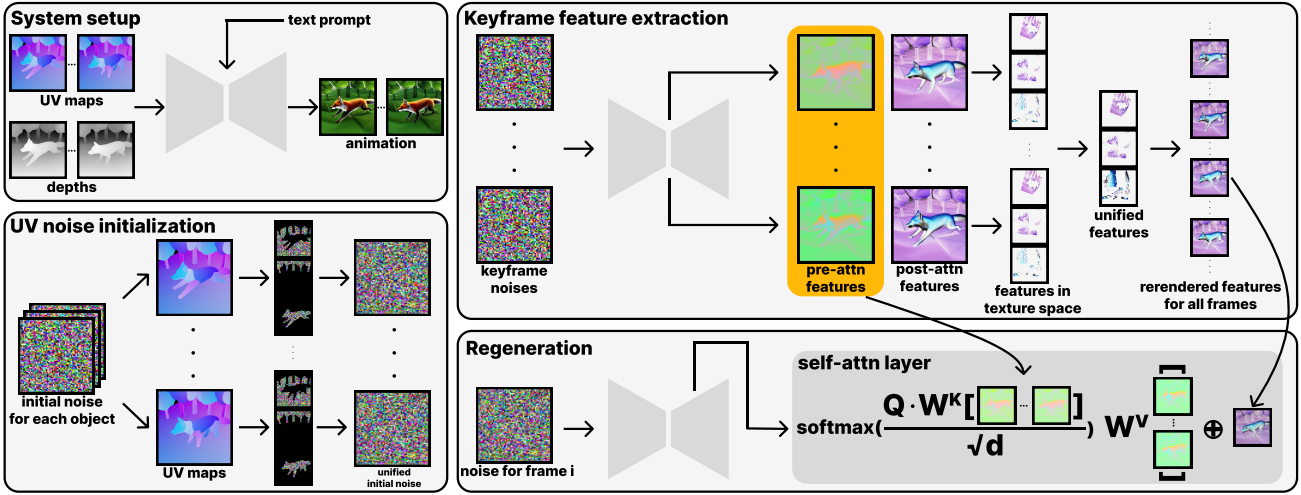}
\end{center}
\vspace{-15pt}
\caption{
\textbf{Overview of our pipeline}. Our system takes as input a set of UV and depths maps rendered from an animated 3D scene. We use a depth-conditioned ControlNet~\cite{zhang2023controlnet} to generate corresponding frames while using the UV correspondences to preserve the consistency. We initialize the noise in the UV space of each object which we then render into each image ~(Sec.~\ref{sec:noise_initialization_model}). For each diffusion step, we first use extended attention for a set of keyframes and extract their pre- and post-attention features~(Sec.~\ref{sec:feature_injection}). The post-attention features are projected to the UV space and unified. Finally, all frames are generated using a weighted combination of the outputs of the extended attention with the pre-attention features of the keyframe, and the UV-composed post-attention features from the keyframes~(Sec.~\ref{sec:uv_feature_injection}).
}
\label{fig:overview}
\end{figure*}
\section{Related Work}
\label{sec:related}

In this section, we discuss the methods most closely related to ours. A comprehensive review of diffusion models for visual computing can be found in~\cite{po2023state}.

\paragraph{Video Generation}
T2V diffusion models extend T2I models by generating dynamic scenes. The majority of works in this area focus on extending pre-trained T2I diffusion models and finetuning on video data~\cite{ge2023pyoco, blattmann2023alignyourlatents, ho2022video, ho2022imagenvideo}.
A typical method of lifting image-based models to videos is to add a time-aware module. This can include 3D convolutions~\cite{blattmann2023alignyourlatents}, or pseudo-3D convolutions factorized over spatial and temporal dimensions to reduce computational costs~\cite{guo2023animatediff, ge2023pyoco, esser2023gen1, singer2022makeavideo,esser2023gen1}. Recently, \cite{guo2023animatediff} has shown very compelling video generation results by training a motion module with such temporal layers on top a frozen StableDiffusion model~\cite{rombach2022latentdiffusion}.  All of these approaches, however, lack fine-grained structural control of the video generation process in a temporally stable manner. %
Moreover, training a high-quality T2V model is computationally demanding and challenging, in part due to the limited sizes of available captioned high-quality video datasets. This is part of the reason why existing T2V models are limited to outputting only a few seconds of animation. For these reasons, we build on T2I diffusion models and directly leverage ground truth correspondences of the input meshes to generate highly expressive and temporally consistent animations of arbitrary length.

\paragraph{Video Editing and Stylization}
Another line of work focuses on video editing and stylization. Given an input video, video editing/stylization aims to change the input video according to a user-defined prompt, while keeping other components harmonious.
Earlier work, such as \cite{bar2022text2live}, decomposes a video to atlases and performs editing directly on these atlases. This approach obtains perfect consistency, but it is only applicable to certain types of scenes and the decomposition is very slow to compute.
Exemplar works based on diffusion models~\cite{ceylan2023pix2video, esser2023gen1, wu2022tuneavideo, khachatryan2023text2videozero, qi2023fatezero} along this path generally rely heavily on extracting temporal information from the input video, e.g., via DDIM inversion~\cite{hubermanspiegelglas2023ddiminversion}, then inflating the self-attention modules to attend to multiple frames, and propagating the edited contents across different frames. These techniques typically overfit to video inputs and are not directly applicable to our setting. For this reason, we adopt these baselines ensuring a fair comparison to their core methods, and demonstrate that our approach outperforms them. In addition, as the temporal prior is either extracted from the input video or learned from video training data, these works generally lack 3D awareness and may not generate 3D-aware animations.
A recent attempt designed a 4D representation that employs a canonical field and deformation field to each frame, such that one can easily edit the video by changing contents on the canonical field which propagates naturally to the other frames~\cite{ouyang2023codef}. However, this representation still has a relatively weak generalization ability and tends to only work well on very short video clips.

In summary, while related, none of these video editing approaches operate in the same setting as our approach, which takes a low-fidelity, untextured rendering of a scene as input to generate a video.

\paragraph{3D Generation}
This class of methods is able to achieve near-perfect 3D consistency by using underlying 3D representations.
One line of work~\cite{hoellein2023text2room, fridman2023scenescape, cai2023diffdreamer} focuses on scene generation using diffusion models. These works typically predict or estimate the scene geometry on the fly using monocular predictors such as MiDaS~\cite{ranftl2022midas} and use that information to constitute individual generations into a scene. This approach is applicable only in specific domains such as indoor scenes or nature scenes, with limited motion control ability due to depth--scale ambiguity.
The closest form of 3D generation to our task is mesh texturing. Given an explicit representation such as a mesh, prior works~\cite{richardson2023texture, wang2023breathing, cao2023texfusion} demonstrate convincing asset mesh texturization using the powerful inpainting function of Stable Diffusion~\cite{rombach2022latentdiffusion}.
Other works~\cite{po2023compositional3d, poole2022dreamfusion} use Score Distillation~\cite{poole2022dreamfusion} to directly optimize a neural radiance field~\cite{mildenhall2020nerf}, but they suffer from quality degradation potentially due to heavy classifier-free guidance~\cite{ho2022cfg}.
Follow-up works~\cite{lin2023magic3d} attempt to combine the strength of score distillation and mesh texturization. %
Since these methods generate a static texture for a given mesh, they are not able to capture appearance interactions (e.g., reflections or shadows) in case of scene-level content and motions.

\section{Method}
The goal of generative rendering is to transfer creator-defined scene-level proxy meshes~(without appearance) and motions directly to a convincing animation. Given a 3D scene composed of animated meshes, we render guiding channels that consist of depth maps $\calb{D}=[\bs{D}^1,...,\bs{D}^N]$ and UV coordinate maps $\calb{UV}=[\bs{UV}^1,...,\bs{UV}^N]$ where $N$ is the length of the sequence and each UV map contains the 2D texture coordinates as well as an object ID in its RGB channels. 
Using a depth-conditioned 2D image generation model such as StableDiffusion~\cite{rombach2022latentdiffusion} with depth ControlNet~\cite{zhang2023controlnet}, we aim to transfer the sequence $(\bs{D}^i,\bs{UV}^i)$ into a set of generated images that depict a stylized animation. To achieve this goal, we perform feature blending over the input and output features of the self-attention modules of the diffusion model (Sec.~\ref{sec:uv_feature_injection}) and also perform UV-guided noise initialization (Sec.~\ref{sec:noise_initialization_model}) to improve the temporal consistency. We provide an overview of our framework in Fig.~\ref{fig:overview} and our overall algorithm in Alg.~\ref{alg:gen_rendering}. Next, we provide some background on the self-attention features of a diffusion model before discussing the different parts of our method in detail.

\subsection{Self-attention feature injection} \label{sec:feature_injection}
Recent work has demonstrated that self-attention features encode spatial self-similarity~\cite{Tumanyan_2023_CVPR} and they are effective in terms of maintaining temporal consistency when they are used to perform cross-frame attention~\cite{ceylan2023pix2video,wu2022tuneavideo}. A typical approach is to extend the diffusion network to process multiple frames jointly where they can attend to each other's features. Furthermore, it is possible to manipulate both the input and output features of a self-attention module to further enforce consistency. We denote these operations as \emph{pre-attention} and \emph{post-attention} feature injections respectively and provide a description in the following.

\paragraph{Pre-attention Feature Injection} 
An attention module is defined as:
\begin{equation}
    \bs{F} = \texttt{Attn}\!\left({\bs{Q};  \bs{K}; \bs{V}} \right) = \texttt{Softmax}\!\left(\frac{\bs{Q} \bs{K}^\top }{\sqrt{d}}\right)\cdot{\bs{V}},
    \label{eq:attn}
\end{equation}
where $\bs{Q}$, $\bs{K}$, and $\bs{V}$ denote the query, key, and value features respectively. In case of a self-attention module, $\bs{Q}$, $\bs{K}$ and $\bs{V}$ are computed by projecting the same features $\bs{f}^{(i,l)}$, via the corresponding attention matrices, i.e.:
\begin{equation}
    \bs{Q}^{(i,l)} = \mathbf{W}^Q \bs{f}^{(i,l)}\!\!, \, \bs{K}^{(i,l)} = \mathbf{W}^K \bs{f}^{(i,l)}\!\!, \, \bs{V}^{(i,l)} = \mathbf{W}^V \bs{f}^{(i,l)}
    \label{eq:QKV}
\end{equation}
where $i$ denotes the frame index and $l$ denotes the diffusion step and $\mathbf{W}^{Q/K/V}$ are weight matrices for query, key, and values, respectively. In other words, for each query feature an attention weight is computed based on the similarity between the query and each key feature. Then, the value features are combined based on this attention weight. 

One can manipulate the input features provided to the self-attention module, specifically the key and value features, which we denote as \emph{pre-attention feature injection}. As discussed by \cite{geyer2023tokenflow}, in case of generating $N$ frames, one naive option is to perform \textit{extended attention} where the features $\bs{f}^{(i,l)}$ of all the frames are concatenated and utilized as key and value pairs:
\begin{equation}
\begin{split}
    \bs{K}^{(i,l)} &= \mathbf{W}^K [\bs{f}^{(1,l)},...,\bs{f}^{(N,l)}], \\
    \bs{V}^{(i,l)} &= \mathbf{W}^V[\bs{f}^{(1,l)},...,\bs{f}^{(N,l)}].
\end{split}
\label{eq:extended_attn}
\end{equation}

Extended attention is a synchronized operation, where all frames attend to features of each other resulting in a more consistent generation. However, generating multiple frames together incurs a high computational cost, which in practice limits the number of frames that can be processed in a sequence. For this reason, we utilize extended attention only for a subset of keyframes and propose mechanisms to propagate the keyframe features to the rest of the sequence, as discussed in the next section.

\paragraph{Post-attention Feature Injection}
While it is possible to manipulate the input of the self-attention layer, another option is to directly manipulate the output, which we denote as \emph{post-attention feature injection}. In particular, given known correspondences between two frames $i$ and $j$, one can re-project $\bs{F}^{(i,l)}$, the output of the self-attention module of frame $i$ at diffusion step $l$, to obtain $\bs{F}^{(j,l)}$:
\begin{equation}
\bs{F}^{(j,l)} = \pi_{i, j}(\bs{F}^{(i,l)}),
\label{eq:post_attn_feature_warping}
\end{equation}
where $\pi_{i, j}$ denotes the re-projection operation, which is enabled by the ground truth pixel-level correspondences between frames given by our UV maps.

While post-attention feature injection is effective in achieving consistency, it tends to create blurry results and also avoids generating of certain interactions between objects (e.g., shadows, reflections) since it overwrites the output the self attention.
Hence, as described next, our method combines pre- and post-attention feature injection in an effective way to generate consistent but also high-quality frames.

\subsection{UV-space feature injection} \label{sec:uv_feature_injection}
Given that pre- and post-attention features of the self-attention module can be manipulated to enhance consistency, our main goal is to perform this manipulation guided by the ground truth correspondences we have from the 3D scene. Specifically, we utilize the canonical UV space to bring the features from the different frames into correspondence which we denote as \emph{UV-space feature blending}. Given a set of feature maps for each frame $i$, we project them to a canonical UV space. In particular, for each texel in the UV space, we find its closest point correspondence in each frame. Having such correspondences from multiple frames, a straightforward approach is to take the (weighted) average to blend the features~\cite{geyer2023tokenflow}. However, we find this consistently leads to sub-optimal blurry results. Therefore, to combat over-smoothing of the features, we additionally perform blending sequentially and fill a certain texel with the features of its corresponding pixel in a frame only if it has not been filled before.
The final unified texture is then the mean of the inpainted texture and the average texture. After features from all frames are blended, we obtain a UV-space feature map which we denote as $\bs{T}^{(i,l)}$. Once such a unified feature map is obtained, we can project it back to each frame to obtain $\bar{f}^{(i,l)}$.

Given the UV-space feature blending mechanism, we next describe how we utilize it during the image generation process. For each diffusion step, we first sample a random batch of $m$ keyframes and perform extended attention on these keyframes, extracting both their pre-attention features $\bs{f}_{KF}^{(i,l)}$ and post-attention features $\bs{F}_{KF}^{(i,l)}$.  We further composite a UV space feature map, $\bs{\bar{T}}_{KF}^{l}$, by blending $\bs{F}_{KF}^{(i,l)}$ as described previously. We then perform the diffusion step on all the frames (in case of keyframes, the diffusion step is simply repeated) in a sequential manner. For each frame, we compose the pre-attention features by concatenating the pre-attention features of the keyframes, $\mathbf{f}_{KF}^{l}=[\bs{f}_{KF}^{(1,l)}, ..., \bs{f}_{KF}^{(m,l)}]$, with the pre-attention features of the current frame $\bs{f}^{(i,l)}$ resulting in:
\begin{equation}
\begin{split}
    \bs{K}^{(i,l)} &= \mathbf{W}^K [\bs{f}^{(i,l)},\mathbf{f}_{KF}^{l}], \\
    \bs{V}^{(i,l)} &= \mathbf{W}^V[\bs{f}^{(i,l)},\mathbf{f}_{KF}^{l}].
\end{split}
\label{eq:pre_attention_blending}
\end{equation}
After executing the self-attention module, we blend the output, i.e., post-attention features of the current frame $\bs{\hat{F}}^{(i,l)}$, with the projection of the blended post-attention features to the current frame $\bs{\bar{F}}^{(i,l)}$. Hence, the updated output of the module becomes:
\begin{equation}
    \bs{F}_{out}^{(i,l)} = \alpha \cdot \bs{\bar{F}}^{(i,l)} + (1-\alpha) \cdot \bs{\hat{F}}^{(i,l)}
\label{eq:final_feature_fusion}
\end{equation}
where $\bs{\hat{F}}^{(i,l)}$ is the output of the inflated self-attention block computed using keys and values in Eq.~\ref{eq:pre_attention_blending}.

\subsection{UV noise initialization} \label{sec:noise_initialization_model}
Prior video editing works~\cite{ceylan2023pix2video, geyer2023tokenflow} generally rely on inversion methods, such as DDIM inversion~\cite{hubermanspiegelglas2023ddiminversion}, to invert each source frame to the noise space which are then used as initialization for editing. %
Since source frames in the input video are coherent, this inversion strategy yields relatively coherent noise patterns that in turn helps with consistent editing. %
However, in our case, since we have an untextured scene, inverting rendered frames that lack appearance does not lead to coherent noisy latents. %
Another option is to keep the noise fixed across all frames. However, this results in severe texture-sticking issues as we demonstrate in the supplementary material. We seek an alternative to capture temporal correlations in initial noise patterns. Inspired by previous work~\cite{kass2011noisefiltering,ge2023pyoco}, we leverage the fact that we have a canonical UV space by initializing the noise in this space by creating a Gaussian noise texture. We then project this noise to each frame by utilizing the frame--UV correspondences. As we illustrate in our evaluations, this significantly boosts the coherency of the generated outputs.

\subsection{Latent normalization} \label{sec:latent_normalization}
We notice a color flickering issue, where the overall color distribution and contrast change between frames, even if the motion is limited. Therefore, we apply an AdaIN operation on the predicted latents to make every frame's overall distribution similar to the first frame. The simple AdaIN operation solves this issue for some samples, but we find that calculating the distribution over all pixels is very sensitive to the motions. Therefore, inspired by ~\cite{alaluf2023crossimage}, we calculate the statistics only on the background pixels, which are easy to separate in our setup.

\begin{algorithm}[ht]
    \caption{Generative Rendering}
    \label{alg:gen_rendering}
    \vspace{4pt}
    \textbf{Input:}
    {\small
    UV maps~$\calb{UV}=[\bs{UV}^1,...,\bs{UV}^N]$; depth maps~$\calb{D}=[\bs{D}^1,...,\bs{D}^N]$; text prompt~$\bs{c}$; diffusion model~$\Phi$; projection from UV to image~$\pi_{\calb{UV}}$;
    $T$ diffusion steps; $k$ keyframes
    }
    \vspace{6pt}
    
    $z_{\calb{UV}} \sim \mathcal{N}(\bm{0},\mathbf{I})~$
    \hfill {\small // Sample noise in UV space}\\
    $\mathbf{z} = \mathbf{z}_{T} = \pi_{\calb{UV}}(z_{\calb{UV}})$\\
    \textbf{For} $t=T,\dots,1$ \textbf{do}
    \begin{algorithmic}
        \STATE $KF=\text{shuffle}(\{i_1,\dots, i_k\}) \;\;$ \hfill \text{\small // Sample keyframes}\\
        \STATE $\bs{f}_{KF}, \bs{F}_{KF} \leftarrow \Phi(\mathbf{z}_{KF}, \calb{D}_{KF}, c)$ \hfill {\small // Grab features from} $\Phi$  \\
        \STATE $\bs{\bar{T}}_{KF} = \text{blend}(\pi_{\calb{UV}}^{-1}(\bs{F}_{KF}))$ \hfill {\small // Blends features in UV space}\\
        \STATE $\mathbf{z}_{t-1}=[\;]$ \\
        \STATE \textbf{For} $z^i$ in $\mathbf{z}$ \textbf{do} \\
        \STATE  \;\; $\bs{\bar{F}}^i = \pi_{UV^i}(\bs{\bar{T}}_{KF})$ \\
        \STATE  \;\; $z^i_{t-1} = \Phi(z^i, \calb{D}_{KF}, c, \bs{f}_{KF}, \bs{\bar{F}}^i) $ \\
        \STATE  \;\; $\mathbf{z}_{t-1}\text{.append}(z^i_{t-1})$ \\
         \STATE  \;\; $\mathbf{z} = \mathbf{z}_{t-1}$ \\

    \end{algorithmic}
    
    \textbf{Output:} $\mathbf{z}$
\end{algorithm}
\begin{figure*}
\begin{center}
\centering
\includegraphics[width=0.99\linewidth]{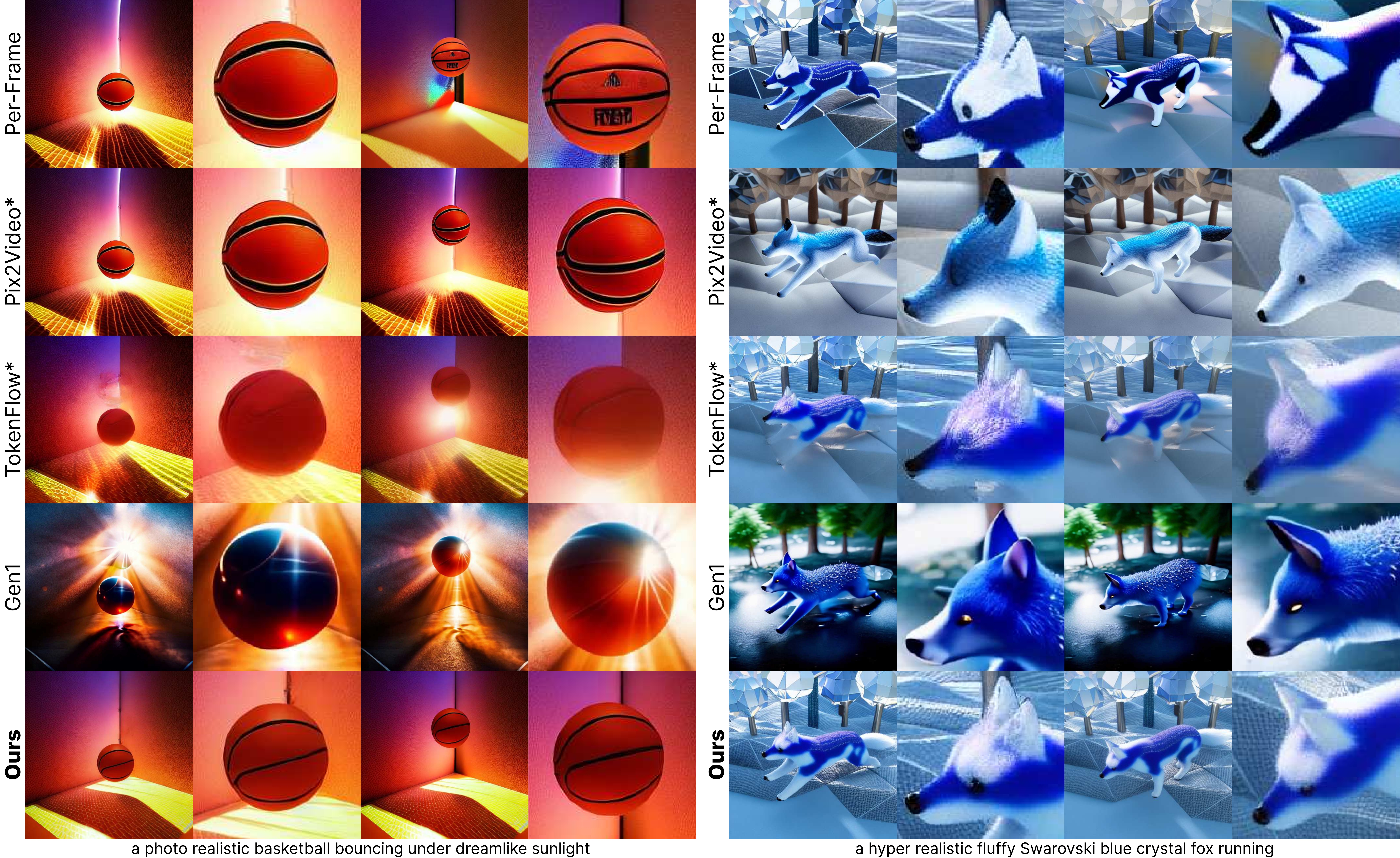}
\end{center}
\vspace{-15pt}
\caption{
\textbf{Qualitative comparisons.} We adapt the feature blending mechanisms proposed by Pix2Video~\cite{ceylan2023pix2video} and TokenFlow~\cite{geyer2023tokenflow} to our setting. While Pix2Video produces reasonable results, it lacks a mechanism to maintain frame-level consistency. TokenFlow uses a token propagation \& interpolation mechanism between nearby keyframes, but this results in blurred outputs in our setting. 
}
\label{fig:qualitative_comparisons}
\end{figure*}
\begin{figure*}
\begin{center}
\centering
\includegraphics[width=0.99\linewidth]{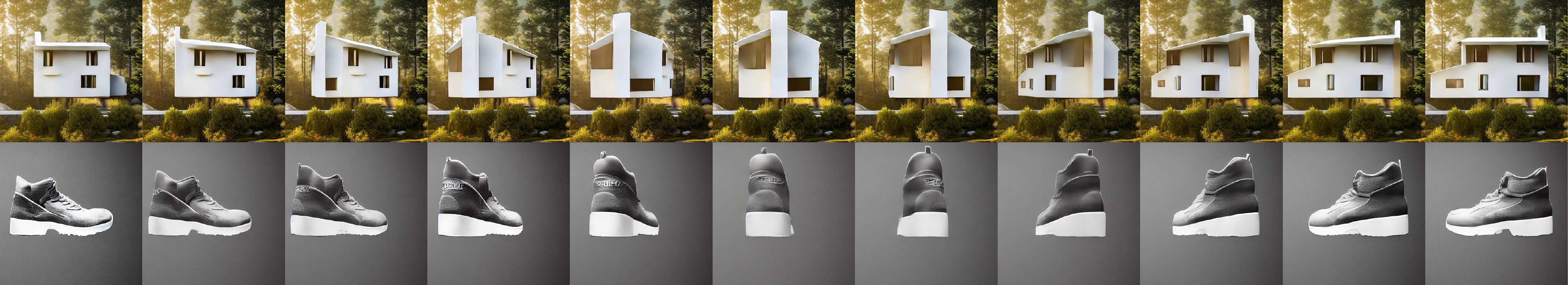}
\end{center}
\vspace{-15pt}
\caption{
\textbf{Camera and object rotation.}
Our algorithm supports camera and object rotations for static scenarios.
}
\label{fig:rotations}
\end{figure*}
\begin{figure*}
\begin{center}
\centering
\includegraphics[width=0.99\linewidth]{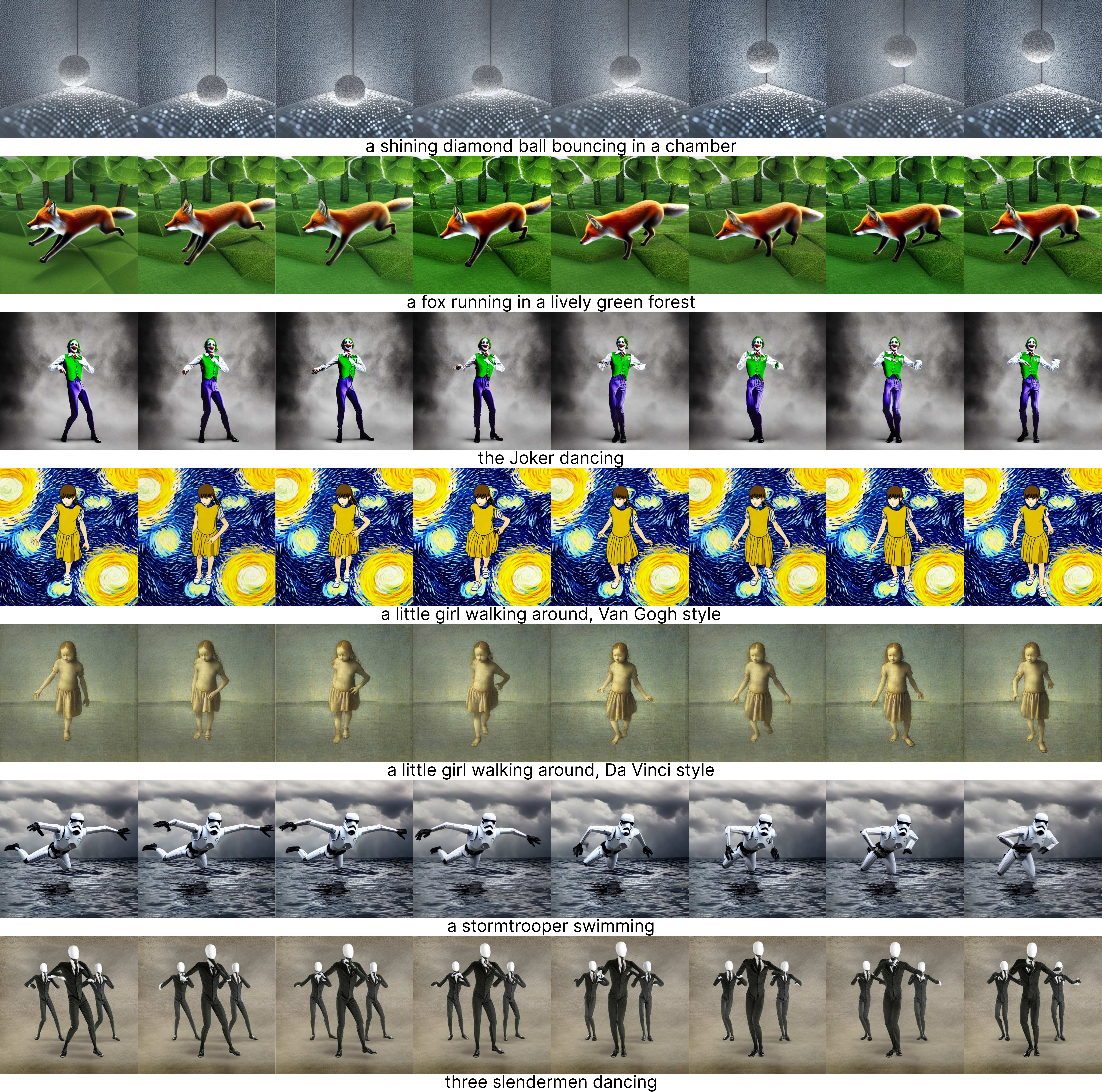}
\end{center}
\vspace{-15pt}
\caption{
\textbf{Qualitative results.} Our method generalizes to a diverse set of input meshes, styles, and prompts.
}
\label{fig:different_realizations}
\end{figure*}
\begin{figure*}
\begin{center}
\centering
\includegraphics[width=0.99\linewidth]{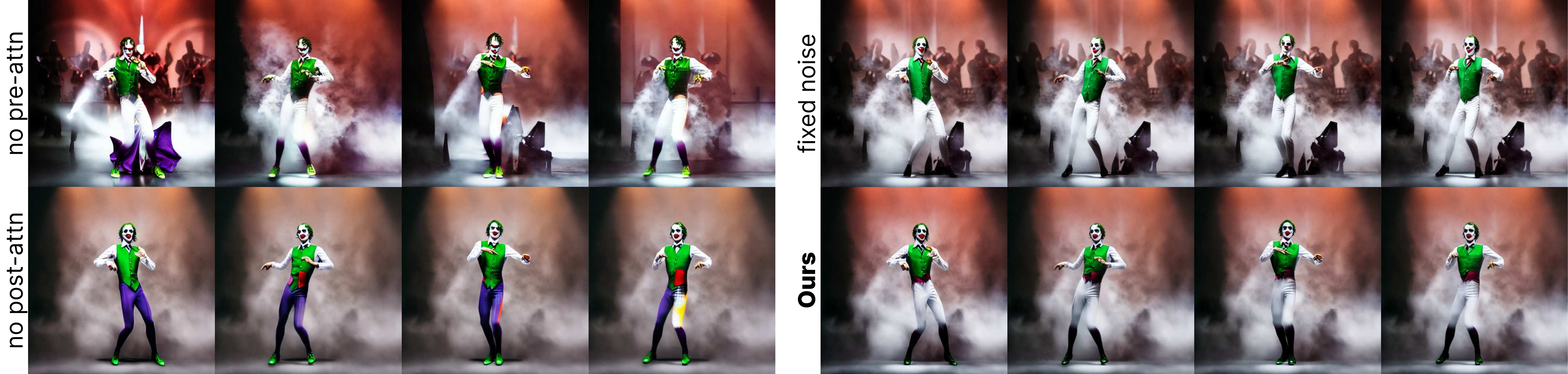}
\end{center}
\vspace{-15pt}
\caption{
\textbf{Ablation study.} We disabled several key building blocks and observe clear artifacts or inconsistencies in all ablations.
}
\label{fig:ablation_study}
\end{figure*}
\section{Experiments}
\label{sec:experiments}
\paragraph{Datasets}
We evaluate our method on scenes with various complexity demonstrating three main types of animation sequences:
\textbf{1)} Camera rotations: given a single static object we create a camera path with 360-rotation around the object;
\textbf{2)} Physical simulation: we prepare a motion sequence by utilizing physically based simulation, e.g., bouncing balls where we use rigid body dynamics for collision handling;
\textbf{3)} Character animation: we create scenes with rigged and animated characters obtained from Mixamo\footnote{https://www.mixamo.com/} and render them with various dance motions.

\paragraph{Baselines}
To our knowledge, no published work directly focuses on the same task our method does. Therefore, we adapt two recent methods to our setting as baselines. 1) we use the pre-attention feature propagation mechanism of Pix2Video~\cite{ceylan2023pix2video} (i.e., each frame attends to an anchor and previous frame) together with depth conditioning, and 2) we apply the extended attention mechanism and token propagation from TokenFlow~\cite{geyer2023tokenflow} using nearest neighbor matching. We note that the original TokenFlow method computes correspondences by matching features obtained via inverting original frames. Since this is not meaningful in our case, we instead provide the ground truth correspondences we obtain from 3D. Both Pix2Video and TokenFlow rely on inversion for noise initialization. Again, this is not meaningful in our setting, so we instead use the same noise to initialize each frame. %

\paragraph{Evaluation metrics}
Following prior video editing evaluation protocols~\cite{blattmann2023alignyourlatents, geyer2023tokenflow, ceylan2023pix2video, esser2023gen1, wu2022tuneavideo}, we report two metrics based on CLIP~\cite{radford2021clip}. To measure \emph{frame consistency}, we compute CLIP image embeddings for each frame of the output video and compute the average cosine similarity between all pairs of consecutive video frames. To measure \emph{prompt fidelity}, we compute the mean CLIP embedding similarity score between each frame of the output video and the corresponding input prompt.

\paragraph{Implementation details}
We use the StableDiffusion~\cite{rombach2022latentdiffusion} v1.5 model and a depth-conditioned ControlNet~\cite{zhang2023controlnet} as our image generator. We generate all results at 512 × 512 resolution using Karras scheduler~\cite{karras2022edm} with 50 denoising steps. 

\subsection{Results}
\paragraph{Qualitative results}
Qualitative comparisons to the baselines are shown in Fig.~\ref{fig:qualitative_comparisons}.
While Pix2Video~\cite{ceylan2023pix2video} produces high-quality frames it shows limited spatio-temporal consistency, because it purely relies on the cross-frame attention mechanism to attend to corresponding features.
TokenFlow~\cite{geyer2023tokenflow} can produce smoother results, but we find that it consistently produces blurry textures. This is in part due to the lack of DDIM inversion~\cite{hubermanspiegelglas2023ddiminversion}, which provides a decent level of feature similarities across frames. In our setting, this DDIM inversion is not meaningful and skipping the self-attention module to directly blend the features of keyframes means averaging less similar features, resulting in blurry appearance. Without accurate DDIM inversion and correct nearest neighbor matching, the blending oversmoothes the features and lead to blurry outputs of low quality.
In addition, unlike nearest neighbor correspondences computed using DDIM inversion, UV correspondences are not guaranteed to be an exact match in the low-dimensional feature space, which is only $64 \times 64$ pixels in our model, especially around object edges. Consequently, a straight blending of the post-attention features also creates blurry results.

Our method can handle camera rotation with reasonable quality in static object-level scenarios, as shown in Fig.~\ref{fig:rotations}. In Fig.~\ref{fig:different_realizations}, we provide additional results on the same input meshes but with different prompts. Our method shows robustness across a diverse set of prompts and styles. Our inflated attention modules can capture motion-specific traits, such as shadows~(Row 3, Row 7), lighting changes~(Row 1), and reasonable deformations that are not modeled by the input low-fidelity meshes~(Row 1 ground regions).

\paragraph{Quantitative results}
\begin{table}
\centering
\begin{tabular}{l c c}
\toprule
Method & Frame Consistency $\uparrow$ & Prompt Fidelity $\uparrow$
\\
\midrule
Per-Frame & 0.9547 & \textbf{0.3233} \\
Pix2Video* & 0.9630 & 0.2983 \\
TokenFlow* & 0.9822 & 0.2737 \\
Gen1      & \textbf{0.9907} & 0.3029 \\
\textbf{Ours} & \underline{0.9845} & \underline{0.3227}
\\
\bottomrule
\end{tabular}
\caption{\textbf{Quantitative evaluation}. 
Our approach offers the best frame consistency among approaches using T2I models. As expected, only the proprietary T2V model Gen1 observes better temporal consistency. 
Our method also shows a higher prompt fidelity than relevant baselines that offer a reasonable amount of temporal consistency. 
*baseline adapted to our setting.}
\label{tab:quantitative}  
\end{table}
We show quantitative evaluations for frame consistency and prompt fidelity in Tab.~\ref{tab:quantitative}. We observe that our frame consistency is better than all baselines using T2I models. As expected, Pix2Video~\cite{ceylan2023pix2video} suffers more from temporal changes due to the lack of a correspondence injection module. Even though TensorFlow's token propagation encourages smoothness over fidelity, our method still shows better consistency. Only the T2V model Gen1 has a better frame consistency, which is expected because it is trained on video data. All current T2V models, including Gen1, can only generate a few frames and are limited in other ways. Our approach is an attempt to overcome this deficiency using 2D T2I models exclusively. Fidelity-wise, we are slightly better than Pix2Video~\cite{ceylan2023pix2video}, because it reuses the set of pre-attention features from previous frames, while we only reuse a portion of the pre-attention features to generate each frame. Since TokenFlow~\cite{geyer2023tokenflow}'s token propagation mechanism introduces over-smoothing effects, its results are blurry for many of the samples, hence achieving lower fidelity scores.
Our generative rendering utilizes the benefits of pre-attention inflation to gain overall global structural and semantic consistency, while also combining with post-attention feature warping as local guidance. Combined with our UV noise initialization, we outperform or are on par with all relevant baselines in the proposed setting. 

\paragraph{Ablation studies}
\label{sec:ablation_studies}
We perform a thorough ablation study to verify our design choices, shown in Fig.~\ref{fig:ablation_study}. We study three aspects of our pipeline: 1) comparing with pre-attention injection only, 2) comparing with post-attention injection only, and 3) the effectiveness of UV noise initialization.
We observe that directly unifying and projecting the post-attention features without passing through the attention module produces chaotic results~(severe artifacts in ``no pre-attn'', the joker's face and clothes regions). This observation is in line with our hypothesis for the poor performance of the adapted TokenFlow mechanism. When the motion is relatively large, geometrically matched features may not be good matches in the feature space. Therefore, directly warping a feature from one frame to another based on the geometrical correspondence has a high chance of not producing reasonable outputs.
However, when we combine post-attention feature warping with pre-attention feature injection, the post-attention features can act as guidance for the next network layer. Without post-attention feature warping, purely pre-attention feature injection does not guarantee pixel-wise consistency since the attention operation merely computes a correspondence based on feature similarities~(in ``no post-attn'', the joker's clothes change over frames).
Our UV noise initialization is well-aligned with our setup. If we use fixed noise instead, the blending of the features may cause unwanted artifacts~(in ``fixed noise'', the joker's face has blending artifacts and is not recognizable in the third image). Our UV noise initialization serves generally and can work beyond our setup. Additional details on ablations are included in the supplementary.
\section{Discussion}
In this paper, we introduce \textit{Generative Rendering}, a novel zero-shot pipeline based on 2D diffusion models for 4D-conditioned animation generation. Our method can animate creator-defined low-fidelity meshes and motion sequences, thereby bypassing steps requiring significant manual labor such as detailing, texturing, physical simulation, etc. The key idea of generative rendering is to utilize the prior within a depth-conditioned 2D diffusion model to provide the basic structure and physical fidelity to make convincing animations.
This is accomplished by injecting the correspondences into the diffusion model using a combination of pre- and post-attention feature injection while unifying these features in the UV space. Our model demonstrates better frame consistency and prompt fidelity than relevant baselines.

\paragraph{Limitation and future work}
Generative rendering cannot achieve real-time animations due to the multi-step inference of current diffusion models. However, accelerating this inference stage has been an active research area and advances in this area, such as consistency models~\cite{song2023consistencymodels}, can be directly applied to speed up our method.
Furthermore, generative rendering is not yet able to guarantee perfect consistency and preservation of details. This is mainly because our method works purely in the low-dimensional latent space, which is only $64 \times 64$ pixels in our model resulting in imprecise UV correspondences. For the same reason, generative rendering may suffer from unwanted misalignment and artifacts. We believe that applying some of findings to pre-trained video diffusion models to augment them with 3D controllability is an exciting future direction.

\paragraph{Conclusion}
Diffusion models have emerged as state-of-the-art generative methods across multiple domains, such as 2D, 3D, and video generation and editing. generative rendering adds unprecedented levels of control to video generation with T2I diffusion models by bridging the gap between the traditional rendering pipeline and diffusion models.

\paragraph{Acknowledgement}
G.W. was in part supported by Google, Samsung, and Stanford HAI.

{
    \balance
    \small
    \bibliographystyle{ieee_fullname}
    \bibliography{macros,main}
}

\appendix

\twocolumn[
\centering
\large
\textbf{Generative Rendering: Controllable 4D-Guided Video Generation with 2D Diffusion Models} \\
\vspace{0.5em}Supplementary Material \\
\vspace{1.0em}
] %
\appendix

\section{Additional qualitative results}\label{sec:additional_qualitative_results}
We refer to the supplemental webpage for various qualitative results obtained by using different 3D scenes, motions, and target prompts.

\section{Additional quantitative results}\label{sec:additional_quantitative_results}
\begin{table}
\centering
\resizebox{0.99\linewidth}{!}{ %
\begin{tabular}{@{}lccccc@{}}
\toprule
& \multicolumn{5}{c}{Pairwise Frame Interval} \\
\cmidrule{2-6}
Method & 1 & 5 & 10 & 15 & 20 \\
\midrule
Per-Frame & 0.9547 & 0.9173 & 0.9109 & 0.9145 & 0.9062 \\
Pix2Video* & 0.9630 & 0.9503 & 0.9494 & 0.9415 & 0.9471 \\
TokenFlow* & 0.9822 & \underline{0.9754} & \underline{0.9728} & \underline{0.9706} & \underline{0.9712} \\
Gen1 & \textbf{0.9907} & 0.9715 & 0.9582 & 0.9624 & 0.9601 \\
Ours & \underline{0.9845} & \textbf{0.9815} & \textbf{0.9738} & \textbf{0.9749} & \textbf{0.9727} \\
\bottomrule
\end{tabular}
} %
\vspace{-5pt}
\caption{
\textbf{Frame consistency with different intervals.} Our approach offers better or competitive consistencies at different intervals.
*: baseline adapted to our setting.
} %
\label{tab:clip_sim}
\vspace{-10pt}
\end{table}%

We report CLIP~\cite{radford2021clip} embedding similarities for a pair of frames separated by different frame intervals in Tab.~\ref{tab:clip_sim}. As shown, semantic consistency achieved by our method is stable across different intervals. This is potentially because of our texture-based feature aggregation. This provides a canonical representation for the objects that links frames that are temporally further apart even in long sequences. Even though video-based methods such as Gen1~\cite{esser2023gen1} achieve temporally smoother results, we speculate that it is not easy for them to capture such long range interactions.

\section{Robustness to video length}\label{sec:video_length}
Our method can be scaled up to handle long input sequences. To generate a long output video, our method requires the sampling of a sparse set of keyframes, and the unified features in the texture space can be propagated. The only GPU memory bottleneck in our method is the parallel processing of the keyframes using inflated attention. Our algorithm is robust enough to generate long videos with $>$200 frames with $<$12GB GPU memory.

\section{Additional ablations}\label{sec:additional_ablation}
\paragraph{Effectiveness of UV noise initialization}
We find that the performance of many components in our pipeline, e.g. inflated attention, are tightly related to the initial noise. Video editing works~\cite{ceylan2023pix2video, geyer2023tokenflow, yang2023rerender} typically require accurate DDIM inversion~\cite{hubermanspiegelglas2023ddiminversion} as noise initialization. Using DDIM inverted noise encourages temporal coherence and similarities to the input video. However, since our input is textureless UV maps and depths, DDIM inversion constantly fails and will not provide useful information. We instead propose to utilize the canonical UV space to initialize the noise in each frame of the sequence. In our supplemental page, we show the effectiveness of this initialization strategy.

\paragraph{Effectiveness of latent normalization}
Our latent normalization can address a large portion of the color flickering issue, where the overall color distribution shifts randomly for different frames. We notice that small color shifts in the latent space will be inflated by the VAE decoder used in Stable Diffusion~\cite{rombach2022latentdiffusion}, causing the generation process very sensitive to small differences in the latent space. The video comparison is included in our supplemental page.

\section{Limitation and failure cases}\label{sec:limitation_and_failure_cases}
\begin{figure}[H]
\vspace{-15pt}
\begin{center}
\centering
\includegraphics[width=0.99\linewidth]{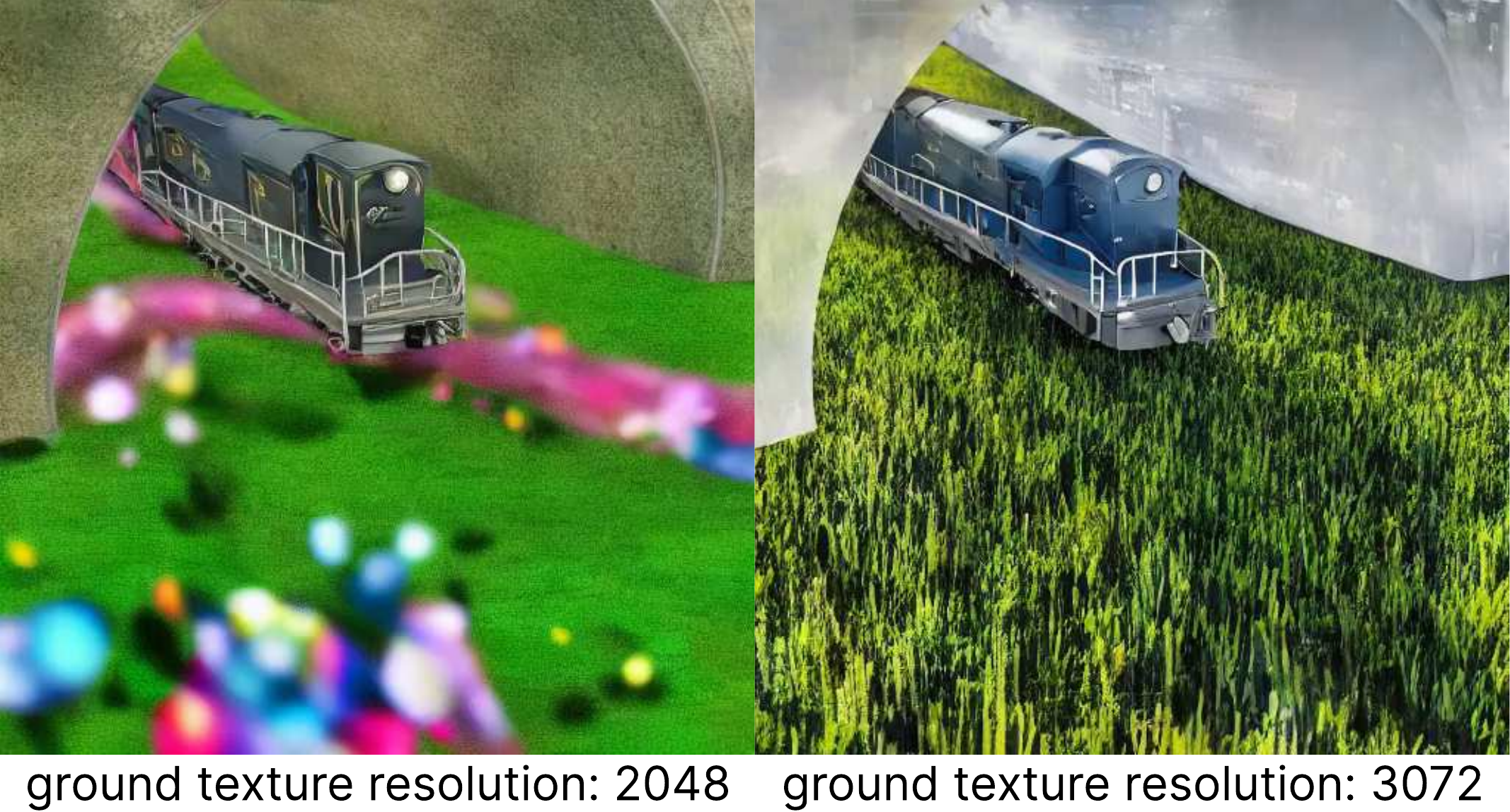}
\end{center}
\vspace{-15pt}
\caption{
\textbf{Effect of different texture resolution.} If the ground's texture resolution is too low~(2048$\times$2048), artifacts appear at ground regions in the final rendered image compared with using a higher texture resolution~(3072$\times$3072). Prompt: a train running in a field of green grass.
}
\label{fig:texture_resolution}
\vspace{-5pt}
\end{figure}
Our method still suffers from several aspects.
A large portion of our inconsistency comes from the VAE decoder. As mentioned in \ref{sec:additional_ablation}, small inconsistencies in the latent space will be inflated by the VAE decoder while transiting into RGB frames. This phenomenon scales to image details, where small differences in the latent space will be inflated after decoding with the VAE.
Additionally, since our noise initialization and feature fusion both work in the UV space, it could be tricky to set the texture resolutions. Setting the texture resolution too low will create corrupted regions while too high will cause the UV coordinates to be too far away and no blending effect will take place. An example is shown in Fig.~\ref{fig:texture_resolution}.
Finally, our work does not yet generalize to large environmental changes and dramatic perspective changes. This is because of the highly overlapping pixel-wise matches and bad feature projections. We believe finetuning or adding a video module to be an exciting future direction to improve the performances on these more challenging scenes.

\end{document}